\documentclass{article}

\usepackage{PRIMEarxiv}

\usepackage[utf8]{inputenc} 
\usepackage[T1]{fontenc}    
\usepackage{hyperref}       
\usepackage{url}            
\usepackage{booktabs}       
\usepackage{amsfonts}       
\usepackage{nicefrac}       
\usepackage{microtype}      
\usepackage{lipsum}
\usepackage{graphicx}
\graphicspath{{media/}}     

\usepackage{subfigure}  
\usepackage{bbding}
\usepackage{diagbox}
\usepackage{amsmath,amsthm}
\newtheorem{theorem}{Theorem}
\usepackage{float}
\title{The Evolution of the Interplay Between Input Distributions and Linear Regions in Networks

}

\author{
  Xuan Qi \\
  \texttt{qixuanxd@163.com} \\
   \And
  Yi Wei \\
  Nanjing University \\
  Nanjing\\
  \texttt{yui.wyown@gmail.com} \\
}

\begin{document}
\maketitle

\begin{abstract}
It is commonly recognized that the expressiveness of deep neural networks is contingent upon a range of factors, encompassing their depth, width, and other relevant considerations. Currently, the practical performance of the majority of deep neural networks remains uncertain. For ReLU (Rectified Linear Unit) networks with piecewise linear activations, the number of linear convex regions serves as a natural metric to gauge the network's expressivity. In this paper, we count the number of linear convex regions in deep neural networks based on ReLU. In particular, we prove that for any one-dimensional input, there exists a minimum threshold for the number of neurons required to express it. We also empirically observe that for the same network, intricate inputs hinder its capacity to express linear regions. Furthermore, we unveil the iterative refinement process of decision boundaries in ReLU networks during training. We aspire for our research to serve as an inspiration for network optimization endeavors and aids in the exploration and analysis of the behaviors exhibited by deep networks.
\end{abstract}

\keywords{Deep neural networks \and Input distributions \and Linear regions \and Piecewise
linear activation functions}

\section{Introduction}
\label{1}
In recent years, due to advancements in DNN (Deep Neural Network) structures such as ChannelNets \cite{book21}, Ghostnets \cite{book18}, and the development of more effective activation functions such as PReLU \cite{book19} and ReLU6 \cite{book20}, DNNs have achieved remarkable progress in various domains and challenging tasks. Nonetheless, the majority of current networks face the challenge of limited interpretability, making it arduous to comprehend the representation of the input space within these networks. Thus, it becomes imperative to delve into the underlying reasons behind the exceptional performance of networks on various classification tasks. An intuitive research method is to transform networks into the mapping of the input space. DNNs fit a variety of different linear functions by using piecewise linear activation functions (e.g. ReLU). As for the standard networks based on piecewise linear activation, it can transform the input space into different linear regions \cite{book1} \cite{book2}. Therefore, when the input distribution is two-dimensional, we can obtain a complete correspondence among the linear regions, the input distribution, and the activation states. In fact, networks output classification results by effectively filling these linear regions with linear functions (based on multi-classification). 

\begin{figure}[htp]
	\centering
	\subfigure[]{\includegraphics[angle=0,width=0.325\textwidth]{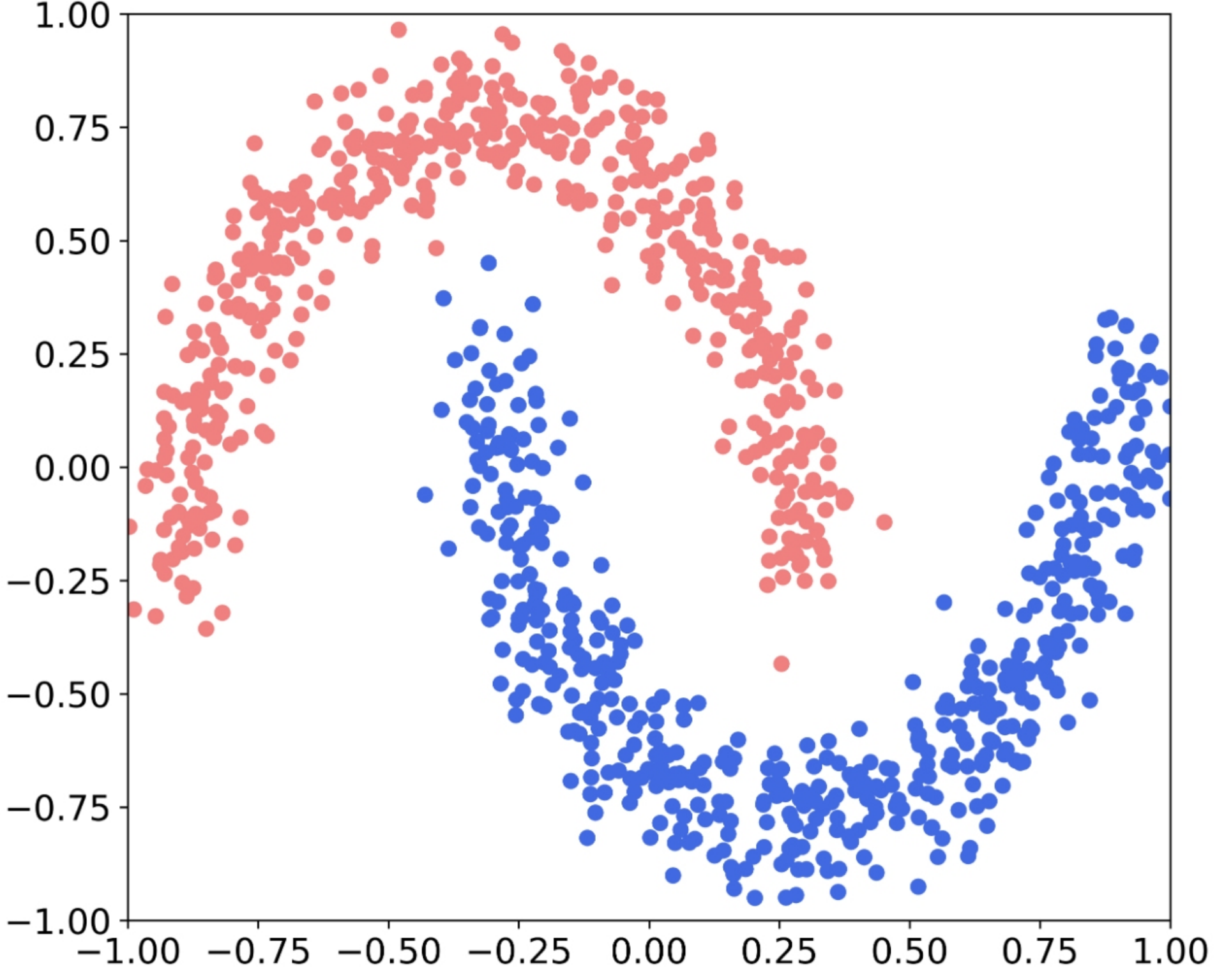}}
	\subfigure[]{\includegraphics[angle=0,width=0.325\textwidth]{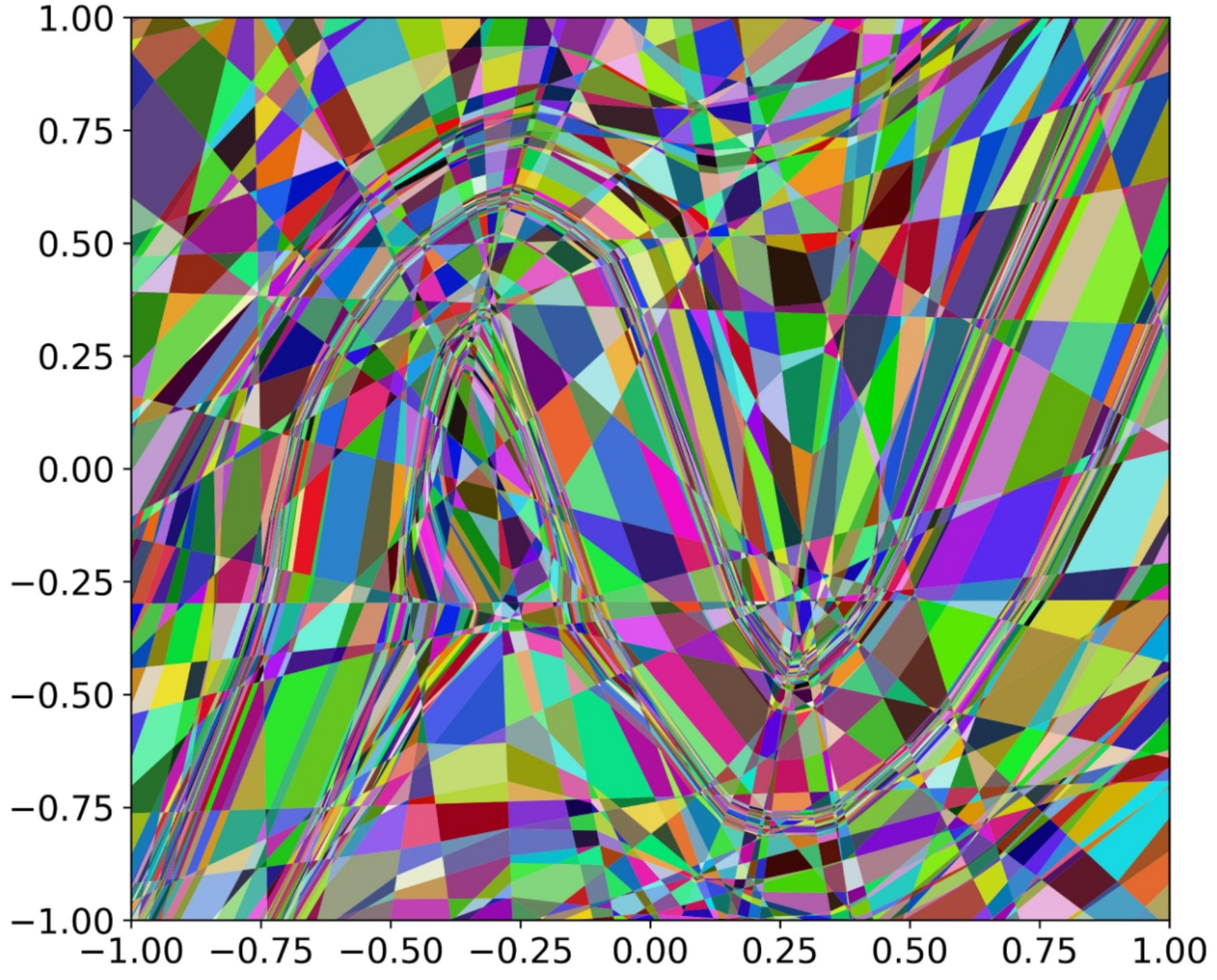}}
	\subfigure[]{\includegraphics[angle=0,width=0.325\textwidth]{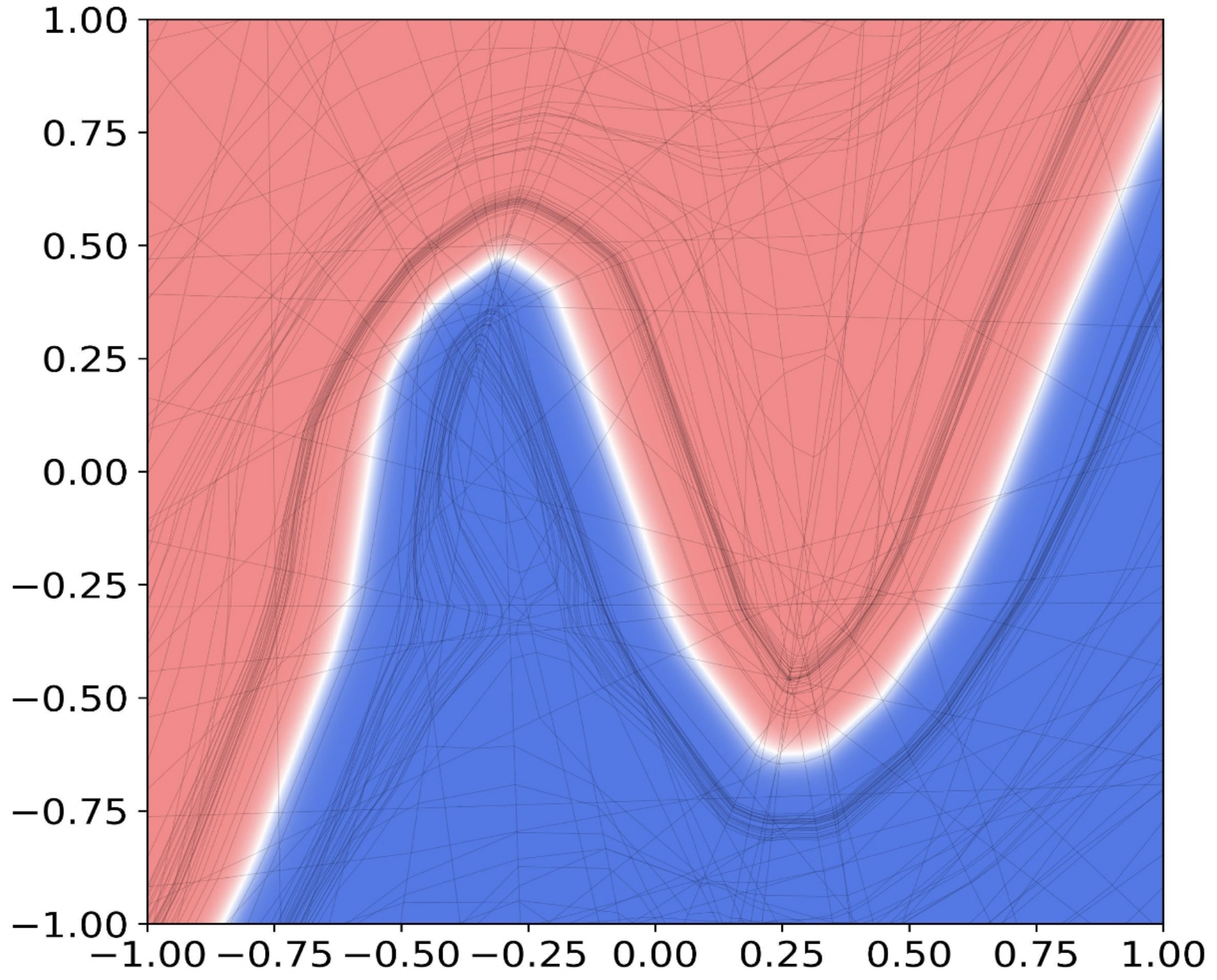}}
	\caption{Where are the linear regions and decision boundaries of DNNs? This figure shows the comprehensive correspondence among the linear regions, the input distribution and the activation states within in a fully-connected ReLU network (3 hidden layers, each with 64 neurons.) functioning in a two-dimensional input space during training for 100 epochs. (a) Two-dimensional input dataset "$make$ $moons$". (b) The visualization of the linear regions, where color blocks represent distinct linear regions of the network. (c) The visualization of the decision regions, where each color represents each type of the input, and the white regions can be regarded as the decision regions approximately.}
	\label{fig:1}
\end{figure}

\begin{figure}[h]
	\centering
	\includegraphics[angle=0,width=1\textwidth]{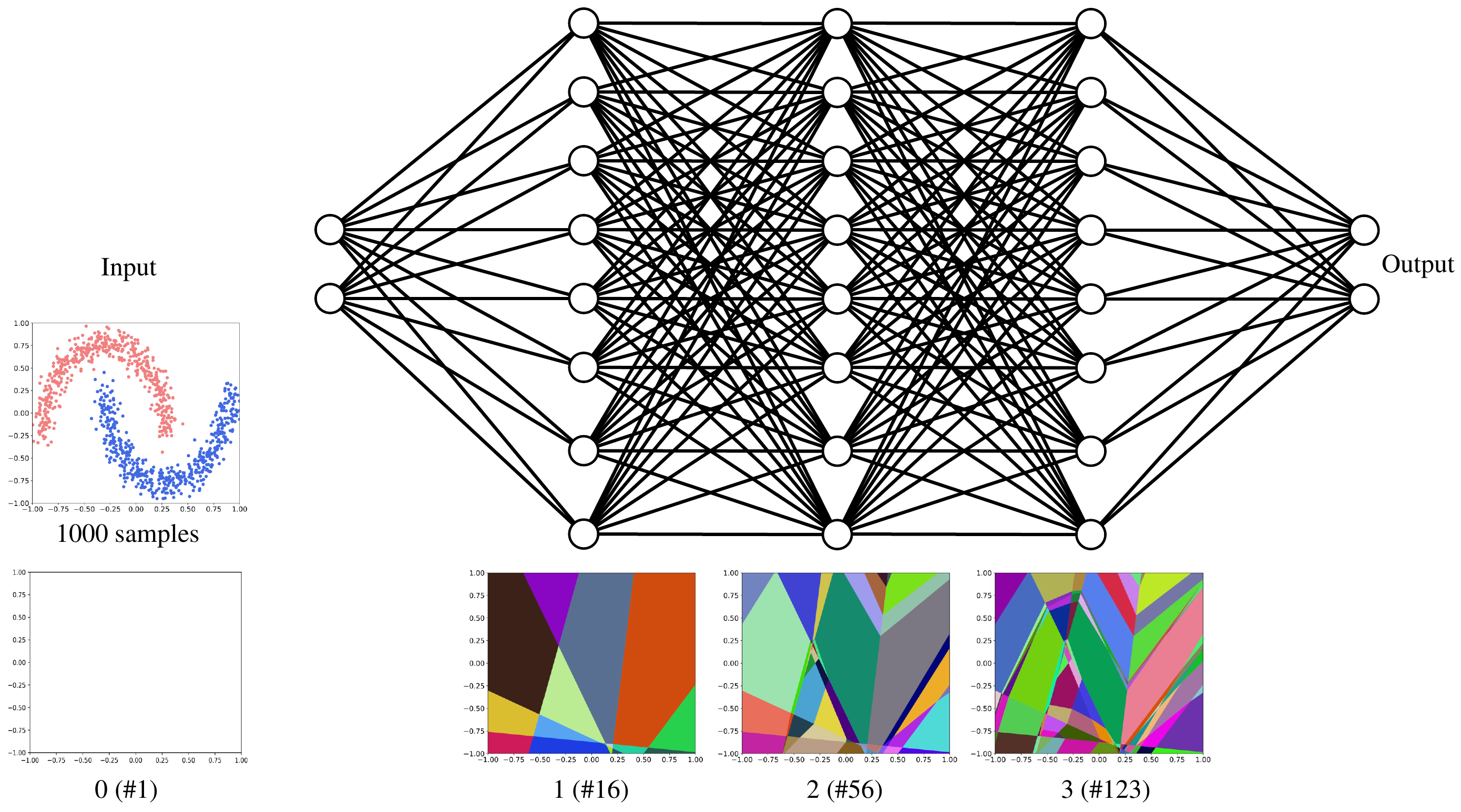}
	\caption{The process of arranging linear convex regions in a fully-connected ReLU network (3 hidden layers, each with 8 neurons.) during training for 100 epochs. We perform a count of the linear regions in each layer. With increasing depth, neurons in each layer combine and segment the linear boundaries inherited from the previous layer, ultimately forming intricate patterns of regions at higher levels. Through this process, the vast number of neurons effectively partitions the input space into numerous distinct regions.}
	\label{fig:2}
\end{figure}

Fig. \ref{fig:1} shows the intricate linear regions of the ReLU network with a two-dimensional input space for binary classification, Each colored block represents a linear region within the neural network. Moreover, each neuron within the ReLU network divides the input space into two regions along the hyperplane. With the collective contribution of numerous neurons, a region can be further divided into multiple subregions. Consequently, each region encompasses two linear functions to express the output result for the binary classification. The prediction label is determined by selecting the maximum value of the result of two linear functions.
As the number of layers increases, the neurons in each layer further partition the regions generated by the previous layer. Fig. \ref{fig:2} shows the progressive arrangement process of linear convex regions in a full-connected ReLU network comprising three hidden layers. Consequently, a network exhibits a multitude of linear regions. These locally dense linear regions serve as a convenient proxy for the local spatial complexity of the network. It should be noted that only a fraction of convex partitions in networks can be characterized as the arrangement of hyperplanes. Therefore, we define the arrangement of linear convex regions in networks based on piecewise linear activation as follows:

Definition 1 Let ${B}_{g}$ be a collection of $G$ linear convex regions, we have
\begin{equation}
	\label{(1)}
	{B}_{g}=\left(B_{1}^{g}, \ldots, B_{d_{g}}^{g}\right),   g \in[G],
\end{equation}
\begin{equation}
	\label{(2)}
	r_{g}=1, \ldots, d_{g},   g \in[G],
\end{equation}
then, the arrangement of these linear convex regions is
\begin{equation}
	\label{(3)}
	{C}\left({B}_{1}, \ldots, {B}_{G}\right),
\end{equation}
and its regions are
\begin{equation}
	\label{(4)}
	\mathbb{E}[\#\{\text { regions in } C\}] = \bigcap_{g=1}^{G} B_{r_{g}}^{g},  (g = 1, \ldots, G).
\end{equation}
In this paper, we conduct a statistical analysis on the number of linear regions in ReLU networks with piecewise linear activation functions. Additionally, we visualize the distributions of these linear regions in a two-dimensional input space. Building upon this, we investigate the relationship between the linear regions and the input distributions in ReLU networks. Our main contributions are summarized as follows:

(1) For a one-dimensional randomly distributed input, there exists a lower bound on the number of neurons required to express it. It's important to note that this lower bound is determined by the number of linear regions that neurons can express, rather than relying on the theoretical value (see Theorem \ref{(32)}).

(2) For the same network, our empirical observations indicate that intricate inputs impede its capacity to express linear regions, and this phenomenon is related to the input complexity exceeding the expressive capacity of the network.

(3) We reveal the evolutionary progression of decision boundaries within ReLU networks during the training process. Furthermore, Our empirical findings suggest that ReLU networks undergo continuous optimization of the decision boundaries during training.

Overall, our findings emphasize the memory capacity of networks operating in a two-dimensional input space when presented with inputs of varying complexity. Furthermore, we partially elucidate the relationship between the operation of ReLU networks and the distribution of inputs. The remainder of this article is organized as follows. Section \ref{2} gives some researchers analysis of the linear regions and operation mechanism of networks. In section \ref{3}, we introduce the linear regions of ReLU networks and informally state our theoretical results. Our detailed experimental process is in section \ref{4}. In particular, Appendix \ref{apd:first} provides the proof of Theorem \ref{(32)}, while Appendix \ref{apd:second} presents our empirical findings.

\section{Related work}
\label{2}
In recent years, a group of researchers has made significant contributions to the analysis of linear regions and the operational mechanisms of networks. The following presents their important contributions in this domain.

\subsection{Expressiveness of depth}
Regarding the influence of network depth on performance, a study conducted by \cite{book4} has shown that for feedforward ReLU networks, changing the depth of the network has a much more significant impact compared to changing the network width, with the impact potentially being exponential. Additionally, \cite{book6} analyzed the necessary and sufficient complexity of ReLU networks in depth. In most cases, networks are universal approximators. It has been demonstrated in \cite{book7} that deep networks are more efficient than shallow ones. Furthermore, \cite{book8} established an upper and lower bound model that can show the complexity of networks in order to study the deep and shallow performance.

\subsection{linear regions of networks}
In relation to the linear regions of networks, \cite{book9} proposed a method to train ReLU networks, enabling the parameters to converge to the global optimum. By partially solving the algebraic topology problem based on ReLU networks, \cite{book26} approximately analyzed how initialization affects the number of linear regions. Furthermore, \cite{book28} conducted calculations and imposed limitations on linear convex regions to analyze the performance of ReLU networks. Additionally, the works of \cite{book27} and \cite{book3} delved into studying and establishing bounds on the maximum quantity of linear convex regions generated by ReLU networks.

\subsection{Analysis of deep and shallow layers}
Several research studies have examined the impact of deep and shallow layers in networks. For instance, the work by \cite{book12} delved into the analysis of why networks with a large number of parameters tend to demonstrate superior generalization performance compared to networks with fewer parameters. In a different study, \cite{book13} proposed scale normalization methods for networks and investigated the generalization performance of ReLU networks. Taking a physics perspective, \cite{book14} explored properties such as symmetry and locality in networks to understand the relationship between deep and shallow layers. Moreover, \cite{book15} suggested that conventional methods fail to explain why networks with more parameters typically achieve better generalization performance than those with fewer parameters. Additionally, \cite{book25} proposed that shallow sub-networks in ReLU networks have priority in training, and partially explained the operational mechanisms of residual blocks.

\section{Linear regions of ReLU networks}
\label{3}
This section introduces the properties and definition of linear regions in ReLU networks. Additionally, we introduce informally that there exists a lower bound on the number of neurons that can express any fixed one-dimensional curve in any high-dimensional input, see Theorem \ref{(32)}.
\subsection{How to think about linear regions}
The ReLU activation function is defined as follows:
\begin{equation}
	\label{(5)}
	ReLU(x)=\max (0, x),
\end{equation}
when the input $x$ is greater than 0, the output of ReLU function is $x$. Conversely, when $x$ is less than or equal to 0, the output is 0 and the corresponding node is inactive and does not contribute to the output. During gradient back propagation, the derivative of ReLU is:
\begin{equation}
	\label{(6)}
	ReLU^{\prime}(x)=\left\{\begin{array}{ll}
		1, & \text { if } x>0, \\
		0, & \text { otherwise }.
	\end{array}\right.
\end{equation}
If $x$ is greater than 0, the gradient equals 1; otherwise, the gradient will vanish. Let’s consider a fully-connected network with $M$ ReLU layers, and the output of each neuron in each ReLU layer is
\begin{equation}
	\label{(7)}
	h^{m}=\left\{h_{i}^{m}, i \in I^{m}\right\},
\end{equation}
where $I^{m}$ is the number of neurons with $m$ layers and $m$ is the number of ReLU layers.
\begin{equation}
	\label{(8)}
	h_{i}^{m}=\left(\mathbf{w}_{i}^{m}\right)^{T} \hat{h}^{m-1}+\mathbf{b}^{m}, \quad i \in I^{m}, m \leq M,
\end{equation}
\begin{equation}
	\label{(9)}
	\hat{h}^{m-1}=\max \left(0, h^{m-1}\right),
\end{equation}
where (\ref{(9)}) is the result of the previous layer passing through the ReLU activation layer, with $\mathbf{b}^{m}$ denoting the bias term of the layer $m$, and $\mathbf{w}_{i}^{m}$ signifying the weight of the neurons in the layer $m$. Therefore, ReLU networks can be expressed as:
\begin{equation}
	\label{(10)}
	f(x)=\max \left(0, \mathbf{w}_{M}^{T} \max \left(0, \mathbf{w}_{M-1}^{T} \max (0, \ldots)+\mathbf{b}^{M-1}\right)+\mathbf{b}^{M}\right),
\end{equation}
referring to (\ref{(10)}), ReLU networks can be regarded as a connected piecewise linear function. Thus, each neuron in the neural network can be expressed as a linear function of input $x \in \mathbb{R}^{n_{i n}}$:
\begin{equation}
	\label{(11)}
	f_{i}^{m}(x)=\left(\mathbf{A}_{i}^{m}\right)^{T} x+\mathbf{b}_{i}^{m}, \quad i \in I^{m},
\end{equation}
where ${n_{i n}}$ is the dimension of the input, $\mathbf{A}_{i}^{m}$ and $\mathbf{b}_{i}^{m}$ are the linear weight and bias of the current neuron input.

Therefore, prior to counting the linear regions of a ReLU layer, we establish the input-based linear expression of neurons within the layer.  Each linear region is defined by the characteristics of the ReLU activation function, and each linear inequality forms a convex region.

We define the linear regions of ReLU networks as follows:

Definition 2 (adapted from \cite{book11}) Let $L$ be a network with input dimension ${v_{i n}}$ and fix $\eta$, the trainable parameter vector of $L$. Define
\begin{equation}
	\label{(12)}
	{\Phi}_{{L}}(\eta):=\left\{x \in \mathbb{R}^{v_{\mathrm{in}}} \mid \nabla L(\cdot ; \eta) \text { discontinuity at } x\right\},
\end{equation}
the linear regions of $L$ at $\eta$ are the connected parts of input without ${\Phi}_{L}$:
\begin{equation}
	\label{(13)}
	\text { linear regions }(L, \eta)=\left\{x \mid x \in \mathbb{R}^{v_{\text {in }}}, x \notin {\Phi}_{L}(\eta)\right\},
\end{equation}

\subsection{Bound for expressing one-dimensional inputs}
Consider a ReLU network $L$ consisting of $p$ neurons with both input dimension and output dimension 1, which has a simple universal upper bound:
\begin{equation}
	\label{(14)}
	\max \#\{\text { regions in } L\}  \leq 2^{p},
\end{equation}
where the maximum is over weight values and bias values. The expressiveness of the number of linear regions depends on the width and depth of DNNs. For the influence of depth and width, see \cite{book24} and \cite{book10}.
\begin{theorem}(informal).
	\label{(32)}
Let $L$ be a ReLU network with one-dimensional input and output. We assume that the weights and biases are randomly initialized so that the pre-activation $o(x)$ of each neuron $q$ has a bounded average gradient \cite{book22}
\begin{equation}
	\label{(15)}
	\mathbb{E}[\|\nabla o(x)\|] \leq K, \quad\quad\quad  some \quad  K>0 ,
\end{equation}
(\ref{(15)}) satisfies the independent zero-center weights initialization of the ReLU neural networks with variance as follows:
\begin{equation}
	\label{(16)}
	V=\frac{2}{\text {fan-in}}.
\end{equation}
Then, for each input $S \subset \mathbb{R}$, the subset $s_{i}$ of $S$ satisfies
\begin{equation}
	\label{(17)}
	\left \{ s_{i}\in S\mid s_{min}\le s_{i}\le  s_{max}  \right \}.
\end{equation}
Indeed, $s_{i}$ is essentially subset that belongs to the one-dimensional curve $S$. According to the above, assuming that the minimum number of linear regions expressing $s_{i}$ is $Q$, we obtain the minimum number of neurons $q$ as follows:
\begin{equation}
	\label{(18)}
	q\approx \frac{\mathbb{E}[Q]}{\lvert s_{i} \rvert \cdot c},
\end{equation}
where $c$ is the number of breakpoints, for ReLU DNNs, $c$ is equal to 1. This result is also sufficient to calculate the number of neurons along any fixed one-dimensional curve in any high-dimensional input. Referring to (\ref{(18)}), we augment the quantity of neurons until the generated linear regions are sufficient to represent the one-dimensional input $S$. At this juncture, the quantity of linear regions generated by the network can likewise denote the input $s_{i}$. Therefore, we obtain the set of neurons that can express the input $s_{i}$ as follows:
\begin{equation}
	\label{(19)}
	\left \{ q\in  N^{*} \mid q\ge \frac{\mathbb{E}[Q]}{\lvert s_{i} \rvert \cdot c}\right \}.
\end{equation}
Considering the effect of random weights and biases on each neuron, for any reasonable initialization \cite{book22}, suppose that the $o(x)$ of $q$ satisfies 
\begin{equation}
	\label{(20)}
	\lvert o^{\prime}(x)\rvert=\gamma (1),
\end{equation}
then, $x \mapsto o(x)$ cannot be highly oscillatory over a large part of $q$ with the input of $s_{i}$. Therefore, let the bias be $b_{q}$ and we have
\begin{equation}
	\label{(21)}
	\mathbb{E}[\#\{\text { solutions in } \left \{ o(x)=b_{q} \right \} \}]=Z(1) ,
\end{equation}
where $Z(1)$ is the solution of the equation we expect. Then, we expect each neuron to generate a fixed quantity of additional linear convex regions. Let the total quantity of additional linear convex regions generated by all neurons be $I$, referring to (\ref{(18)}), we obtain the minimum number of neurons that can express the input $s_{i}$ under the influence of random weights and biases as follows:
\begin{equation}
	\label{(22)}
	q\approx \frac{\mathbb{E}[Q]- I}{\lvert s_{i} \rvert \cdot c} ,
\end{equation}
referring to (\ref{(19)}), similarly, we obtain the set of neurons that can express the input $s_{i}$ under the influence of random weights and biases as follows:
\begin{equation}
	\label{(23)}
	\left \{ q\in  N^{*} \mid q\ge \frac{\mathbb{E}[Q]-I}{\lvert s_{i} \rvert \cdot c}\right \}.
\end{equation}
\end{theorem}
Drawing from the experimental insights presented in \cite{book22}, we have obtained analogous outcomes (see Fig. \ref{fig:10}). We have a great desire to extend our results in subsequent work to calculate a lower bound on the quantity of neurons that can express $\mathbb{R}$. Additionally, we intend to extend Theorem \ref{(32)} to spaces of higher dimensions.
\section{Experiments}
\label{4}
The experimental framework utilized in this study is PyTorch. We represent neurons in each layer of the network as linear functions. In this method, inputs are mapped to neurons to obtain a linear function of each neuron based on the input.

The datasets employed in the experiment consist of two-dimensional data, as it facilitates easy visualization. In reference to \cite{book22}, the 784-dimensional MNIST dataset was randomly screened for data of two dimensions to visualize and count the linear regions. However, for high-dimensional data, selecting data of two dimensions cannot express the whole dataset well. As a result, this study does not employ such datasets for experimentation and exploration.

\begin{figure}[htp]
	\centering
	\subfigure[]{\includegraphics[angle=0,width=0.35\textwidth]{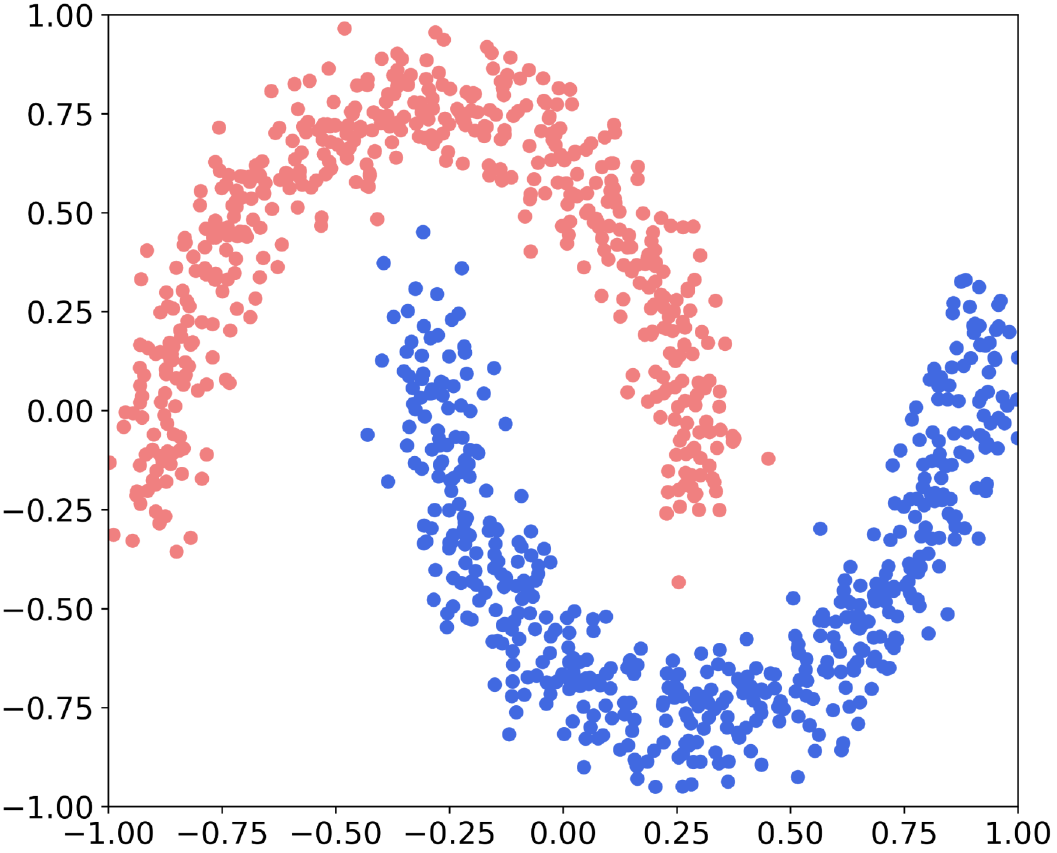}}
	\subfigure[]{\includegraphics[angle=0,width=0.35\textwidth]{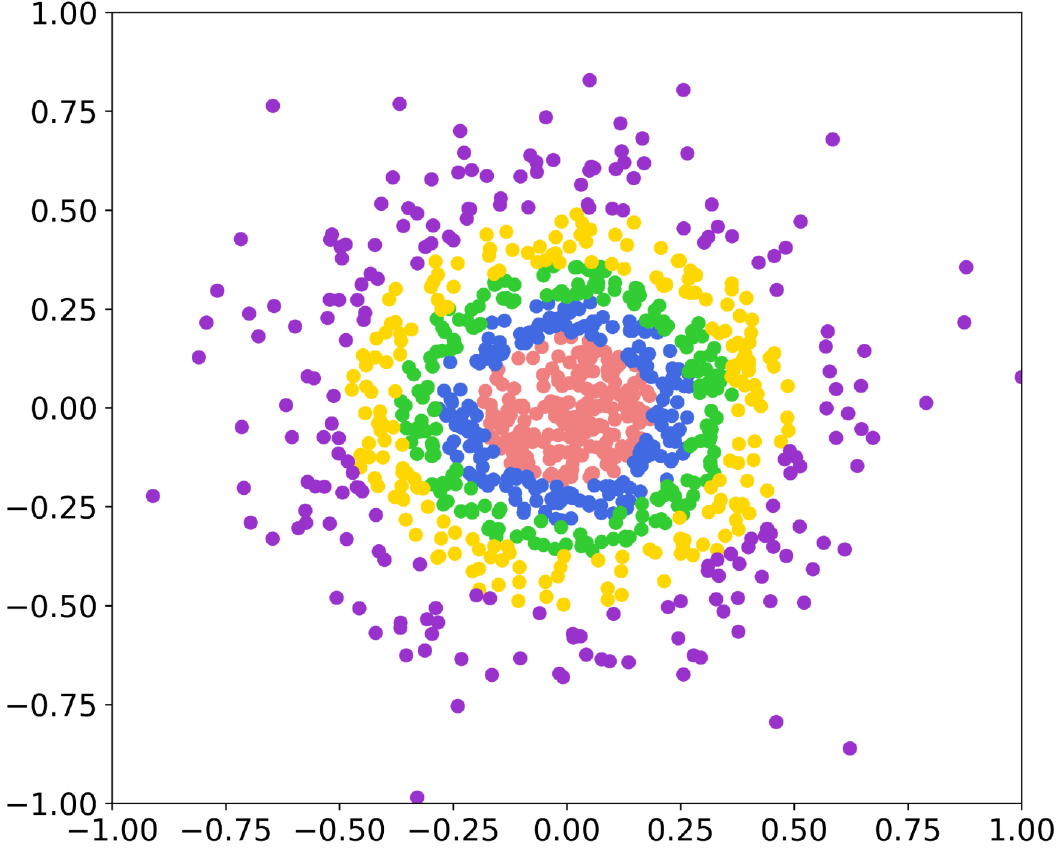}}
	\caption{(a) The "$make$ $moons$" dataset consisting of 1000 samples and 2 classes. (b) The "$make$ $gaussian$ $quantiles$" dataset consisting of 1000 samples and 5 classes.}
	\label{fig:5}
\end{figure}

We use three distinct two-dimensional datasets, with data values ranging from $-1$ to $1$. The first dataset is randomly generated data with different samples evenly distributed between $-1$ and $1$, along with randomly generated labels of either $0$ or $1$. The remaining two datasets are "$make$ $moons$" and "$make$ $gaussian$ $quantiles$". Fig. \ref{fig:5} shows the distribution of "$make$ $moons$" and "$make$ $gaussian$ $quantiles$". In addition, we employ the $Adam$ optimizer, set the batch size to $32$, and utilize the $Cross$ $Entropy$ loss function. In this section, we denote $\left[\mathrm{n}_{1}, \mathrm{n}_{2}, \ldots, \mathrm{n}_{\mathrm{r}}\right]$ to represent the number of hidden layers and neurons, with each hidden layer connected to a ReLU activation layer.
\subsection{Linear regions predicated upon one-dimensional input}
\begin{figure}[htp]
	\centering
	\includegraphics[angle=0,width=0.5\textwidth]{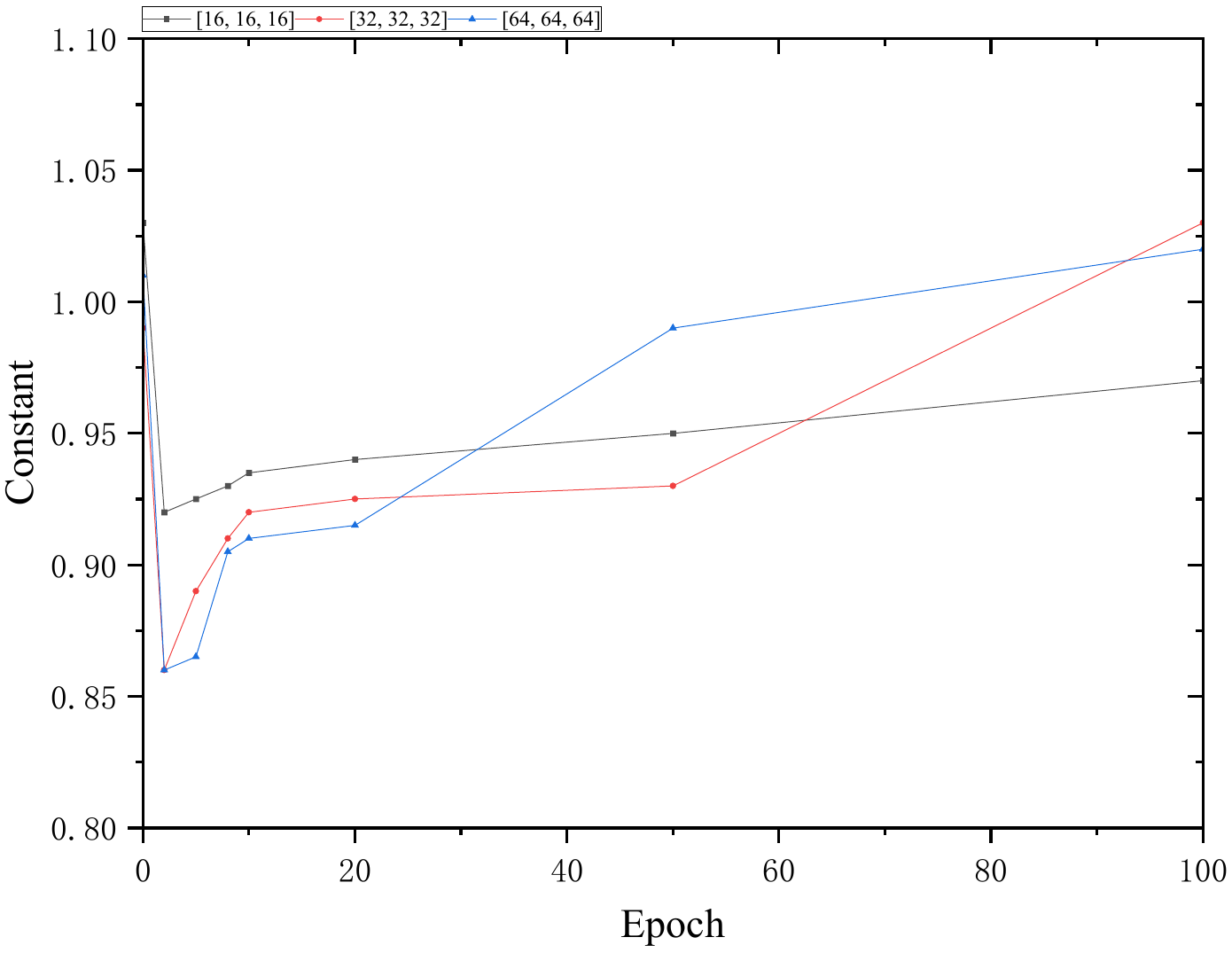}
	\caption{The evolution of the count of linear regions for one-dimensional lines in the input space during the training process of networks $[16, 16, 16]$, $[32, 32, 32]$ and $[64, 64, 64]$ over 100 epochs.}
	\label{fig:10}
\end{figure}

We compute the count of linear regions for the training examples' lines selected randomly from the origin to the input space. Fig. \ref{fig:10} illustrates the average values of networks $[16, 16, 16]$, $[32, 32, 32]$ and $[64, 64, 64]$, each independently trained ten times. The horizontal axis represents the training epochs, while the vertical axis indicates the ratio of the linear regions count to the number of network neurons. We have obtained results closely resembling those presented in \cite{book22}. Namely, throughout the entire training process, the quantity of linear regions does not significantly deviate from the number of neurons in the network, remaining within a constant range. This aligns with the intuitive comprehension described in (\ref{(18)}).
\subsection{Linear regions vary with different inputs}
We use the network with hidden layers $[32, 32, 32]$ and three ReLU layers. Our random samples are $200, 500, 1000, 2000, 5000$ and $10000$ respectively.

\begin{figure}[htp]
	\centering
	\begin{minipage}[t]{0.325\textwidth}
		\centering
		\includegraphics[width=\textwidth]{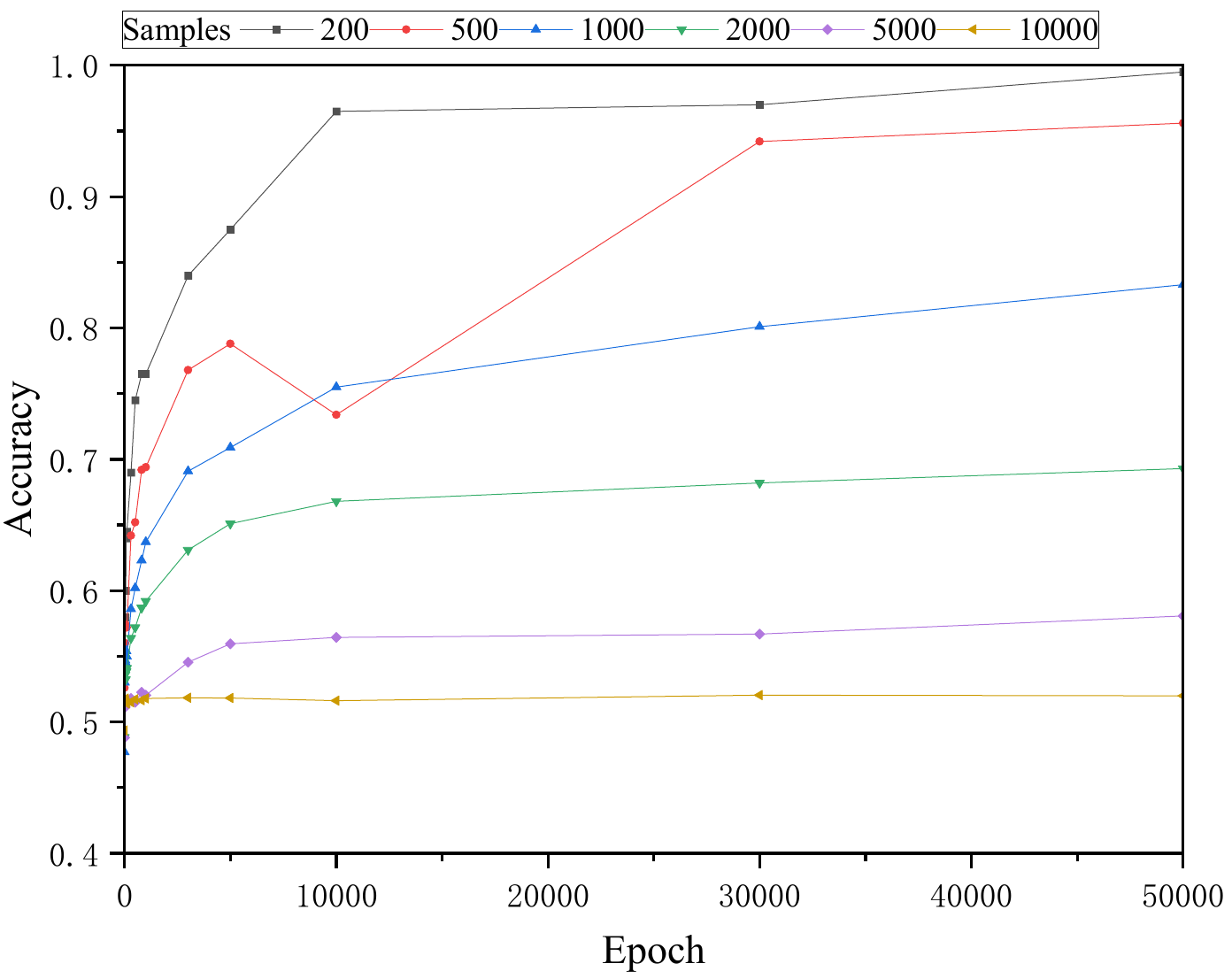}
	\end{minipage}
	\begin{minipage}[t]{0.325\textwidth}
		\centering
		\includegraphics[width=\textwidth]{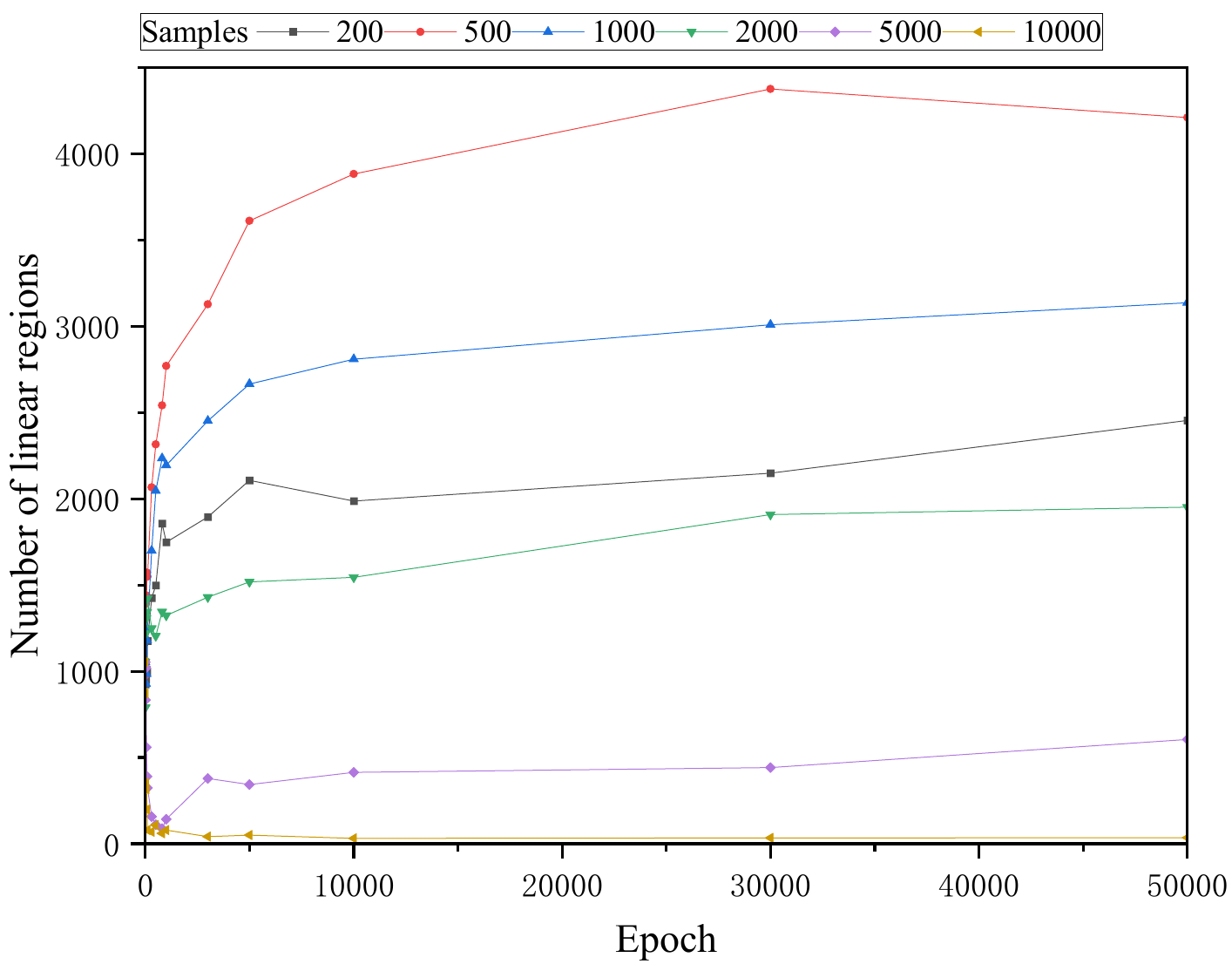}
	\end{minipage}
	\begin{minipage}[t]{0.325\textwidth}
		\centering
		\includegraphics[width=\textwidth]{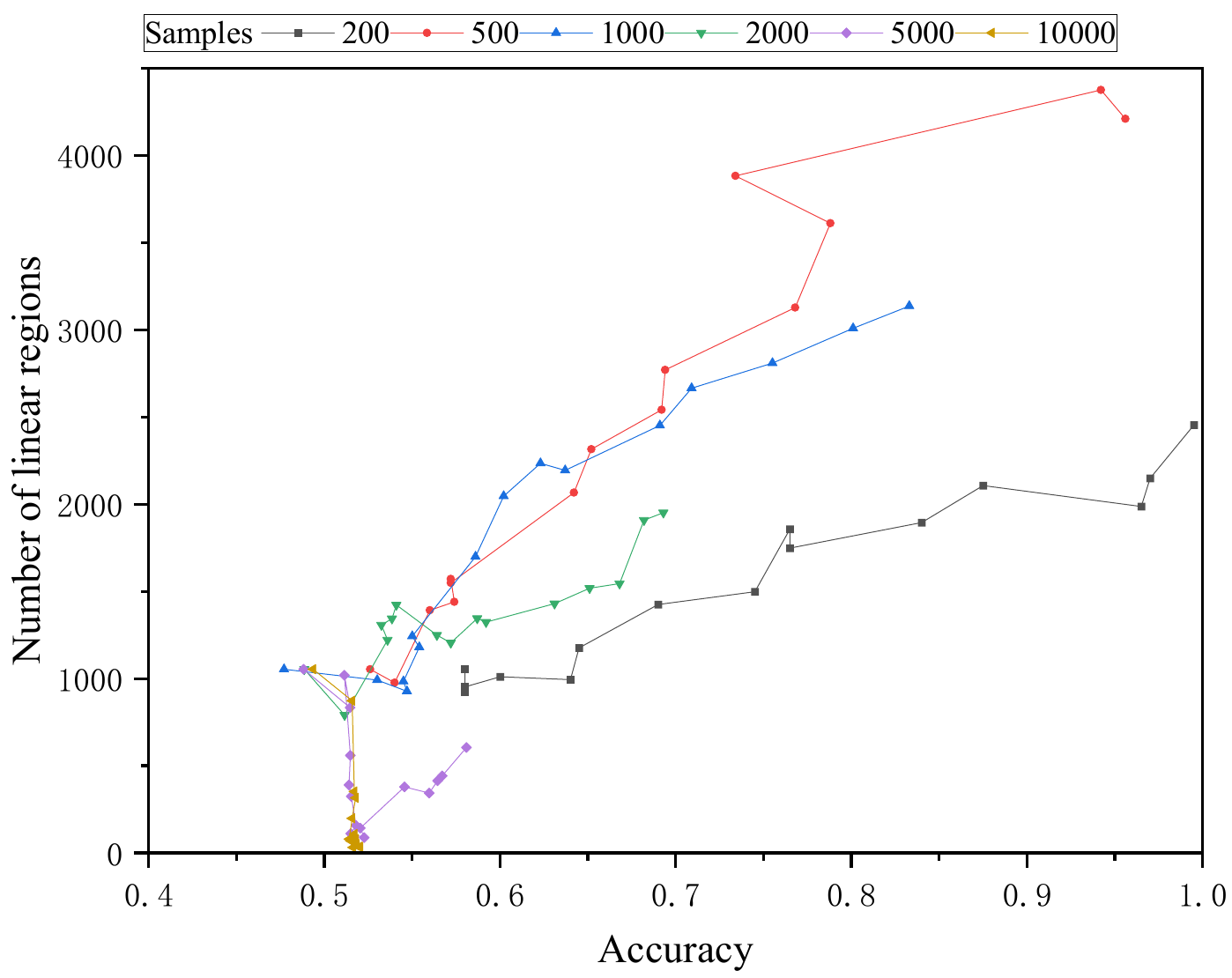}
	\end{minipage}
	\caption{The relationship between the number of linear regions, accuracy, and epochs for random data with varying sample sizes in the network $[32,32,32]$.}
	\label{fig:3}
\end{figure}

\begin{figure}[htp]
	\centering
	\includegraphics[angle=0,width=1\textwidth]{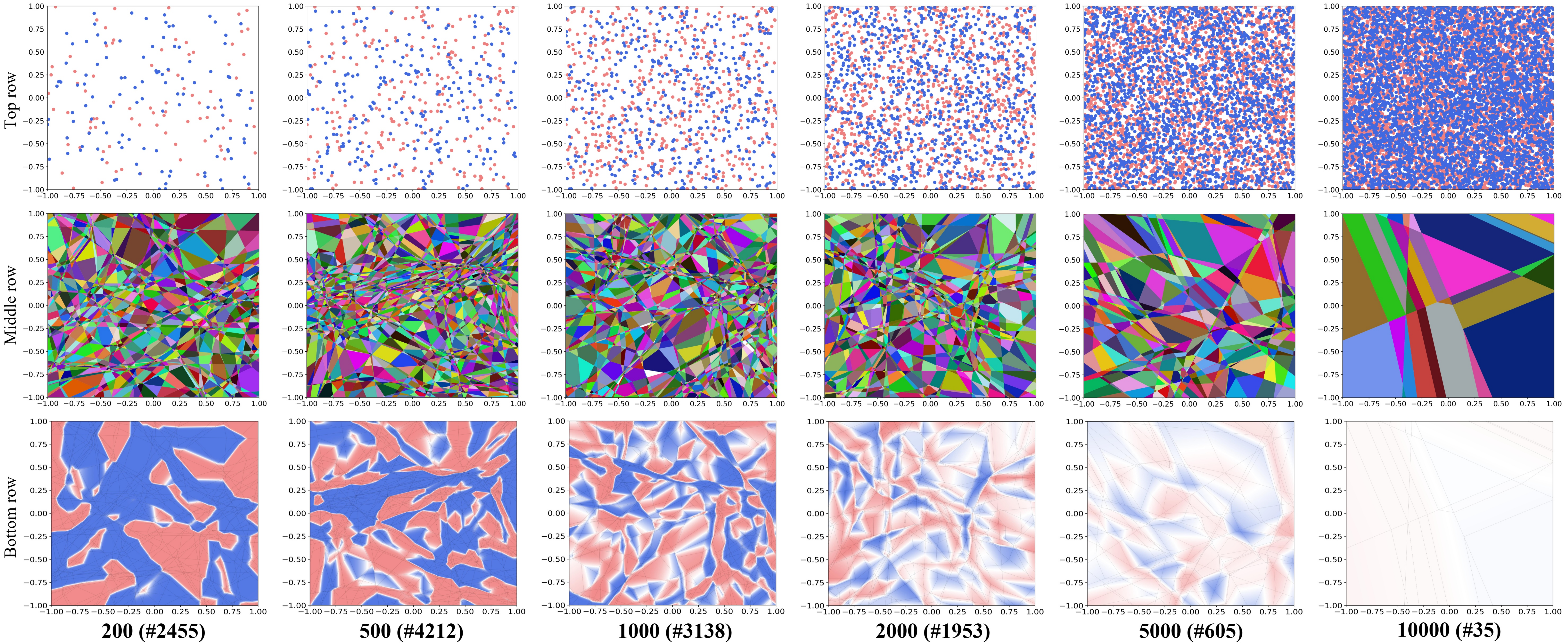}
	\caption{The number of linear regions and the visualization results are presented for random data with varying sample sizes $z$ using the network $[32, 32, 32]$ and training for 50000 epochs, for $z=200, 500, 1000, 2000, 5000, 10000$. Top row: Random inputs used for training. Middle row: Visualization of the linear regions. Bottom row: Visualization of decision boundaries, where each color represents each type of the input data, and the white regions can be approximately regarded as the decision boundaries.}
	\label{fig:4}
\end{figure}

Fig. \ref{fig:3} and Table \ref{table 1} show the results of training for 50000 epochs using different random samples and a learning rate of $0.001$. We can find that as the complexity of the sample space increases, the network is capable of segmenting a larger number of linear regions. Based on our experience, this is due to the need for the model to accommodate more irregular and discrete data by employing additional linear regions. However, when the number of samples breaks through a zero bound point, the linear regions will be greatly reduced as depicted in Fig. \ref{fig:3} and Table \ref{table 1}. When the number of samples reaches $2000$, the accuracy decreases and the number of linear regions experiences a substantial decline. At $10000$ samples, the number of linear regions almost reaches the lowest point. Therefore, our empirical findings suggest that for a network trained on randomly distributed inputs, there exists a critical number of samples that leads to the network's failure in fitting the input. Evidently, this phenomenon is related to the input complexity exceeding the expressive capacity of the network. Fig. \ref{fig:4} shows the linear regions and the visualization results at epoch $50000$ for different samples.

\begin{table}[htp]
	\centering
	\fontsize{9}{15}\selectfont   
	\caption{The number of linear regions in the network [32, 32, 32] for random data with different samples and epochs.}
	\setlength{\tabcolsep}{1.1mm}{
		\begin{tabular}{l|ccccccccccccccc}
			\toprule
			\diagbox [width=5.7em,trim=l] {Samples} {Epoch} & 0 & 10 & 30  & 50 & 80 & 100 & 300 & 500 & 800 & 1000 & 3000 & 5000 & 10000 & 30000 & 50000 \\
			\hline
			200 & 1151 & 925 & 953 & 1011 & 995 & 1176 & 1426 & 1499 & 1858 & 1749 & 1896 & 2108 & 1988 & 2149 & 2455  \\
			500 & 1151 & 977 & 1393 & 1441 & 1572 & 1549 & 2067 & 2316 & 2542 & 2771 & 3129 & 3616 & 3884 & 4377 & 4212 \\
			1000 & 1153 & 992 & 928 & 984 & 1180 & 1244 & 1700 & 2047 & 2236 & 2195 & 2453 & 2667 & 2810 & 3010 & 3138  \\
			2000 & 1154 & 791 & 1222 & 1308 & 1344 & 1423 & 1250 & 1205 & 1346 & 1324 & 1431 & 1519 & 1545 & 1910 & 1953  \\
			5000 & 1151 & 834 & 1020 & 560 & 391 & 325 & 158 & 111 & 88 & 143 & 379 & 344 & 415 & 442 & 605  \\
			10000 & 1152 & 873 & 353 & 316 & 198 & 79 & 70 & 109 & 61 & 80 & 42 & 51 & 31 & 34 & 35  \\
			\bottomrule
	\end{tabular}}\vspace{0cm}
	\label{table 1}
\end{table}

The experimental results also allow us to obtain the following content. Taking the example of 200 samples, during the optimization process, the network undergoes fine-tuning to fit the data.  This fine-tuning process leads to fluctuations in the total number of linear regions within a certain range.
\subsection{The process of fitting decision boundaries}
\begin{figure}[htp]
	\centering
	\begin{minipage}[t]{0.325\textwidth}
		\centering
		\includegraphics[width=\textwidth]{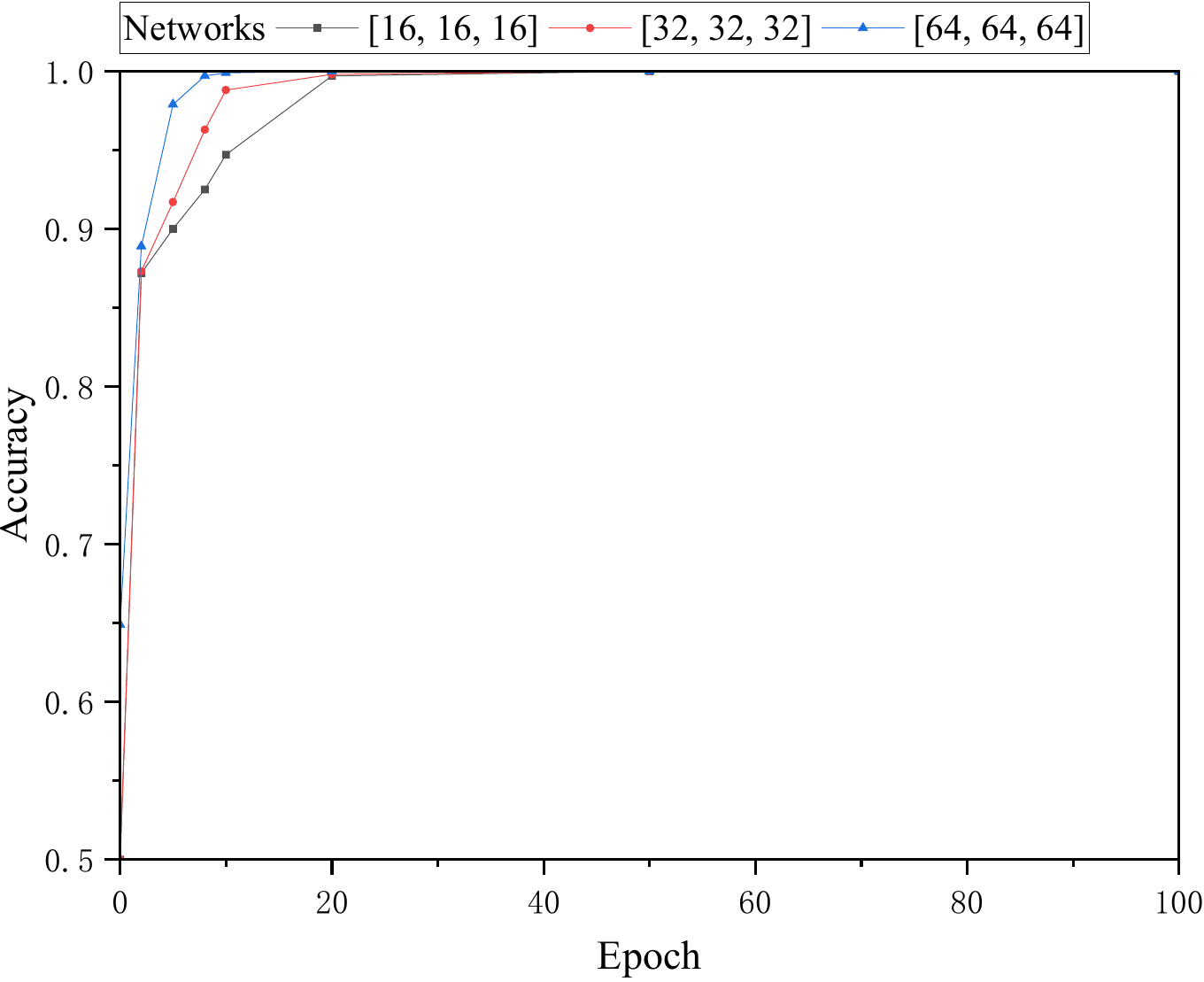}
	\end{minipage}
	\begin{minipage}[t]{0.325\textwidth}
		\centering
		\includegraphics[width=\textwidth]{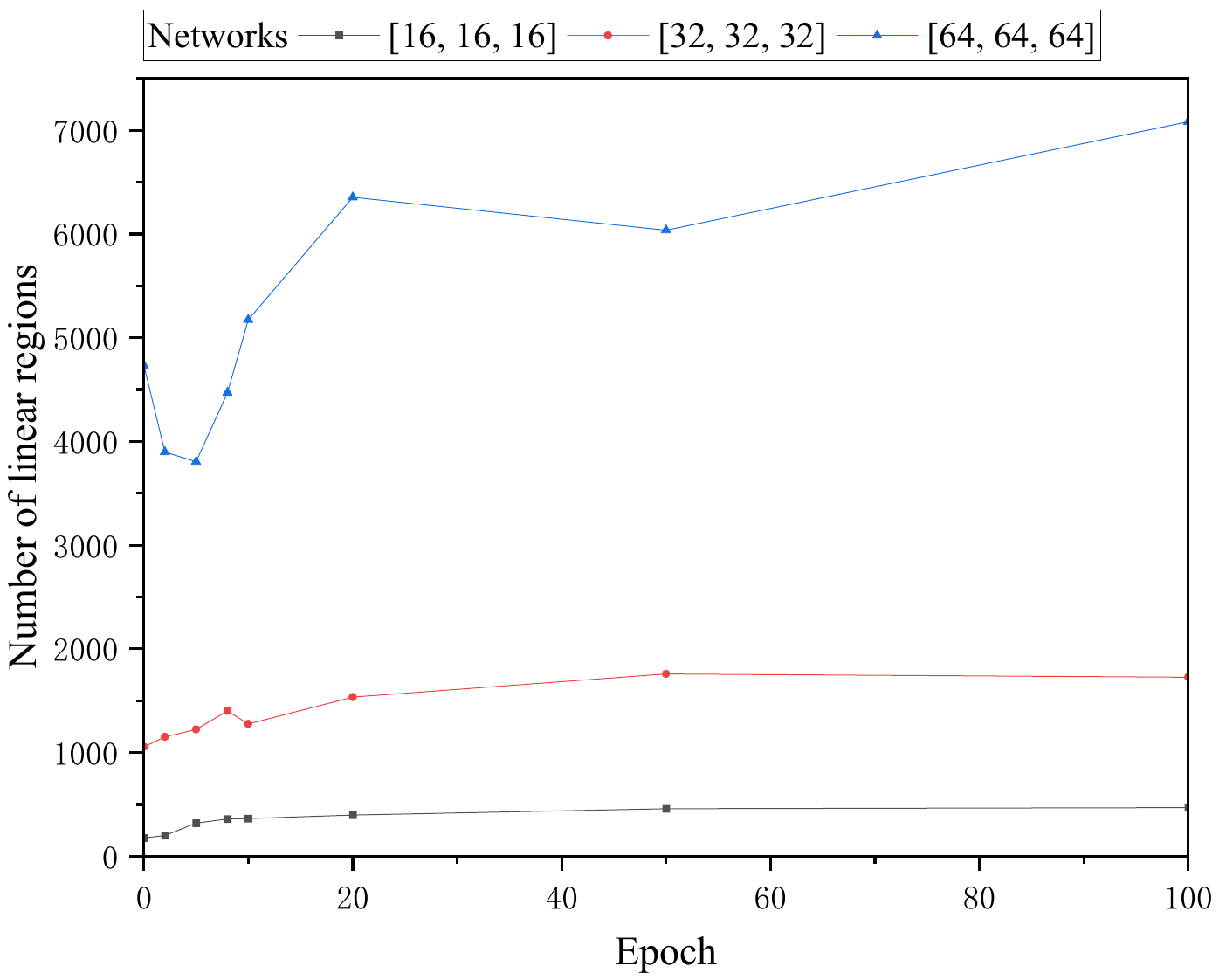}
	\end{minipage}
	\begin{minipage}[t]{0.325\textwidth}
		\centering
		\includegraphics[width=\textwidth]{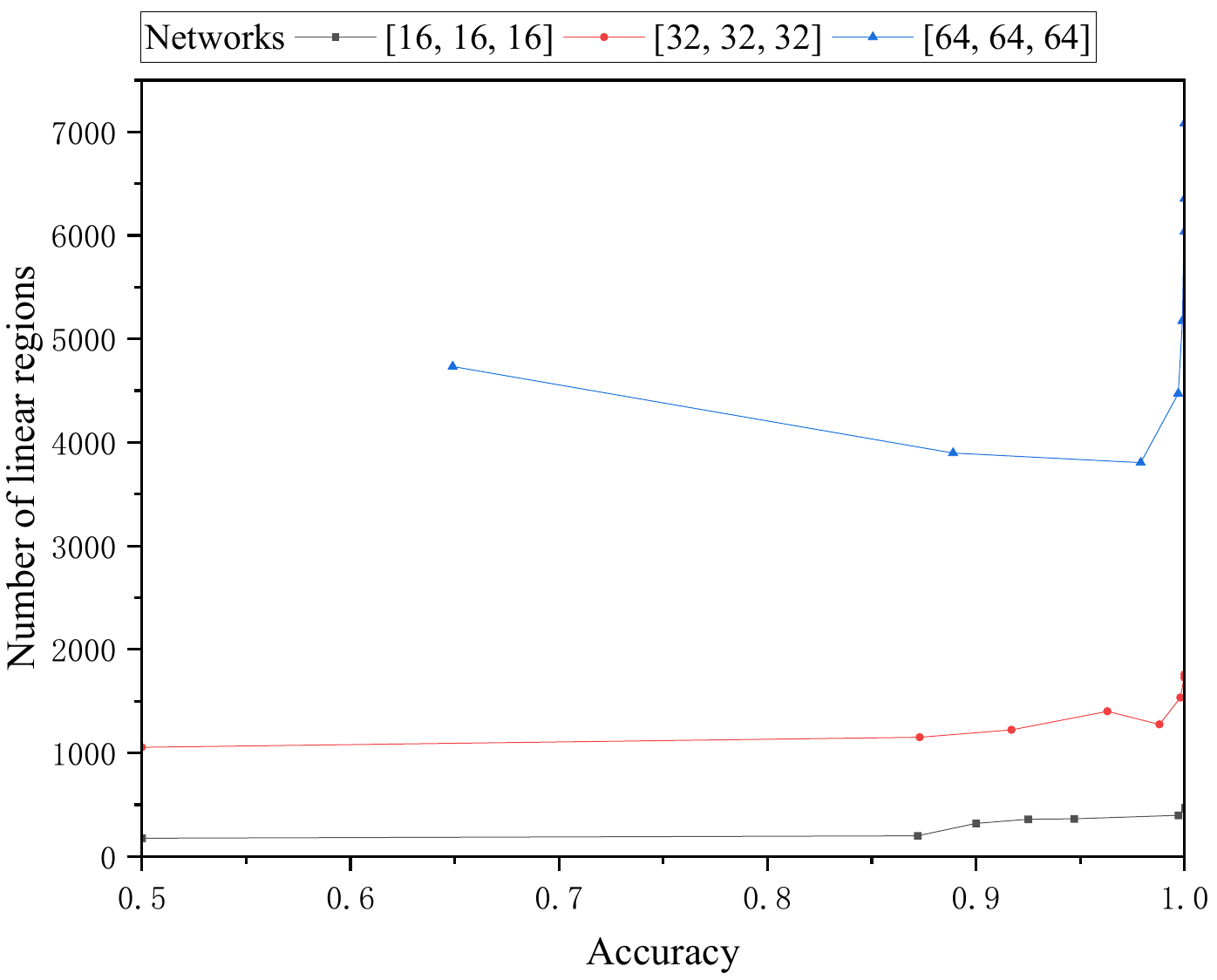}
	\end{minipage}
	\caption{The relationship between the number of linear regions, accuracy and epochs in different networks for the "$make$ $moons$" dataset with 1000 samples.}
	\label{fig:6}
\end{figure}

\begin{figure}[htp]
	\centering
	\begin{minipage}[t]{0.325\textwidth}
		\centering
		\includegraphics[width=\textwidth]{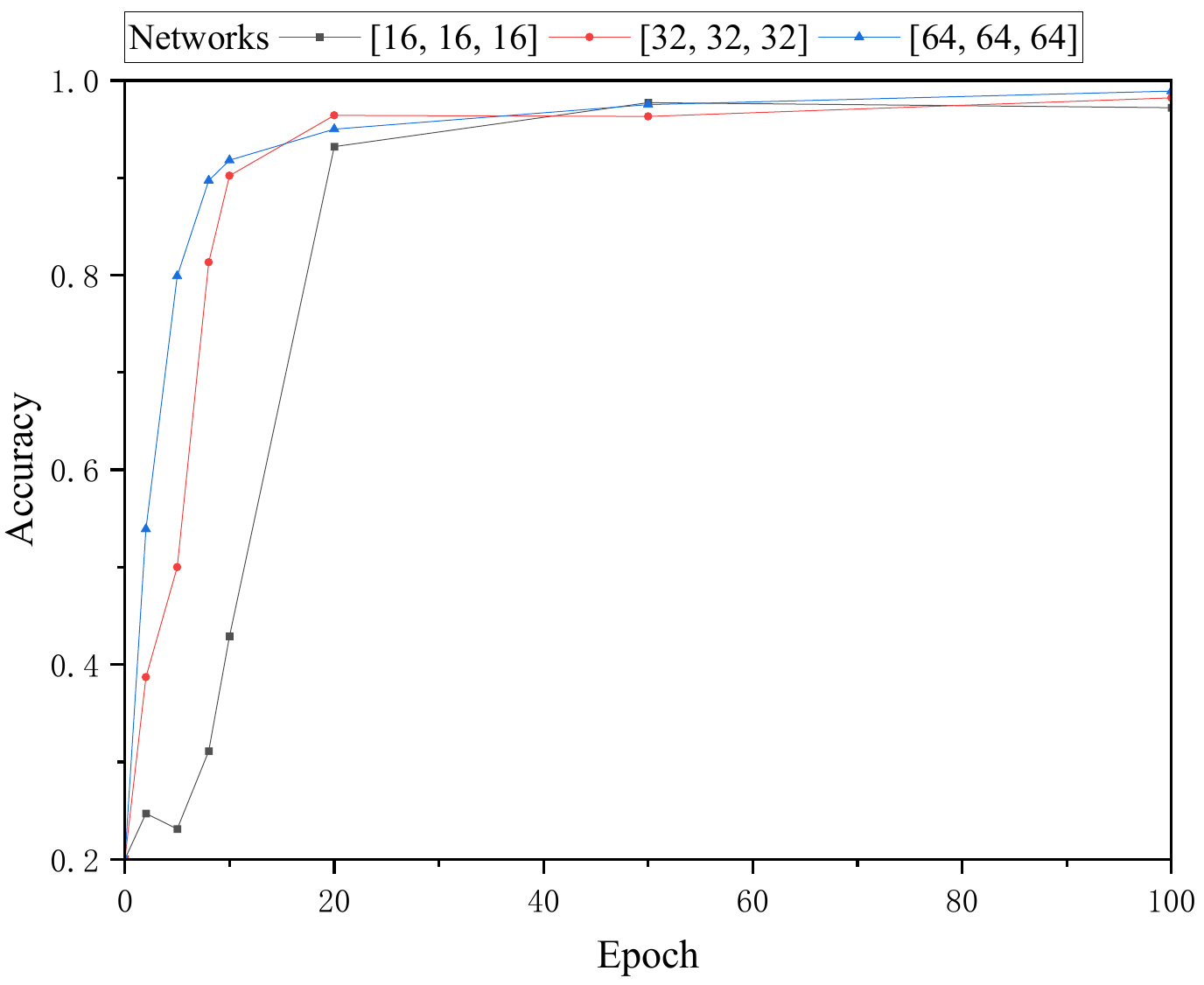}
	\end{minipage}
	\begin{minipage}[t]{0.325\textwidth}
		\centering
		\includegraphics[width=\textwidth]{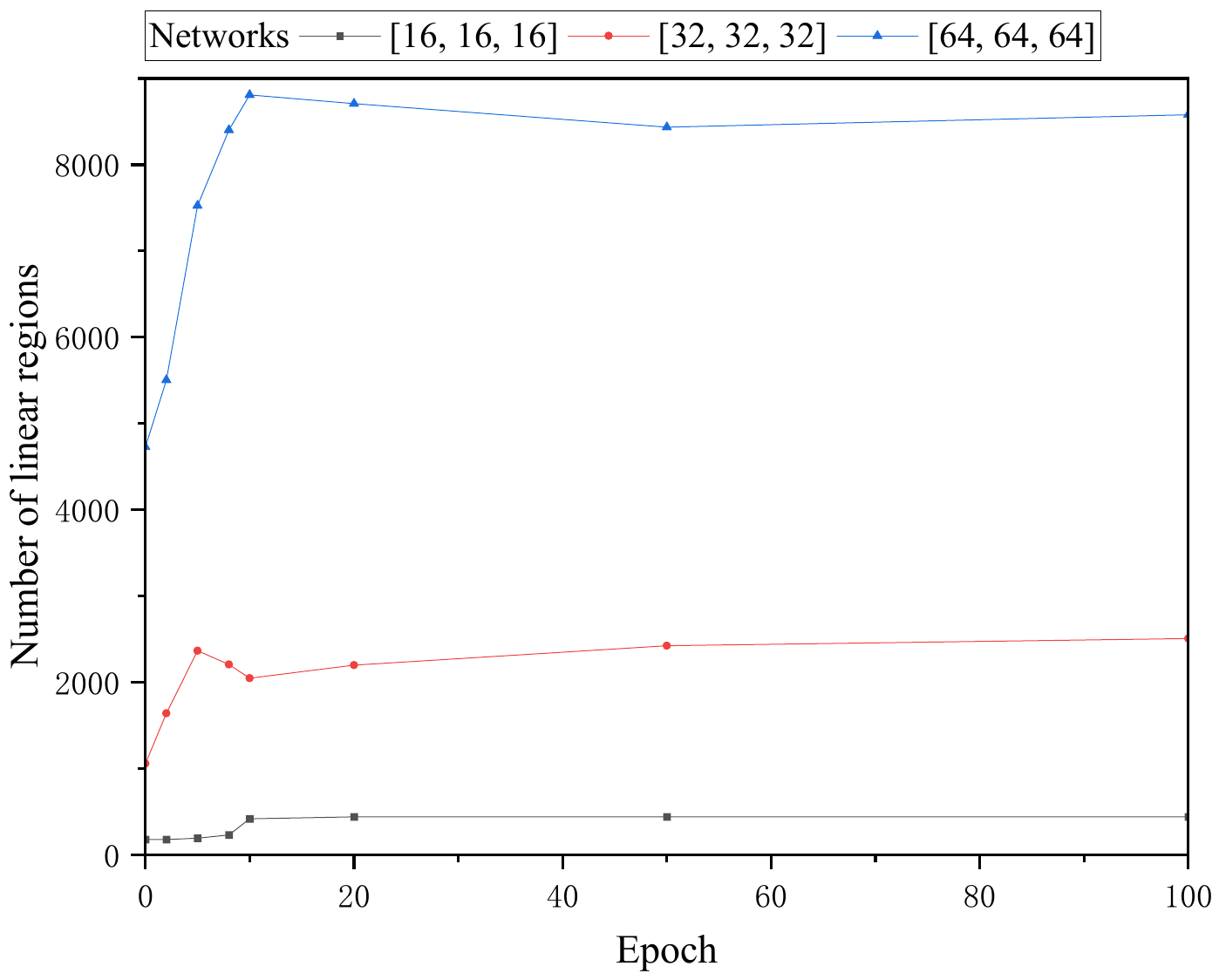}
	\end{minipage}
	\begin{minipage}[t]{0.325\textwidth}
		\centering
		\includegraphics[width=\textwidth]{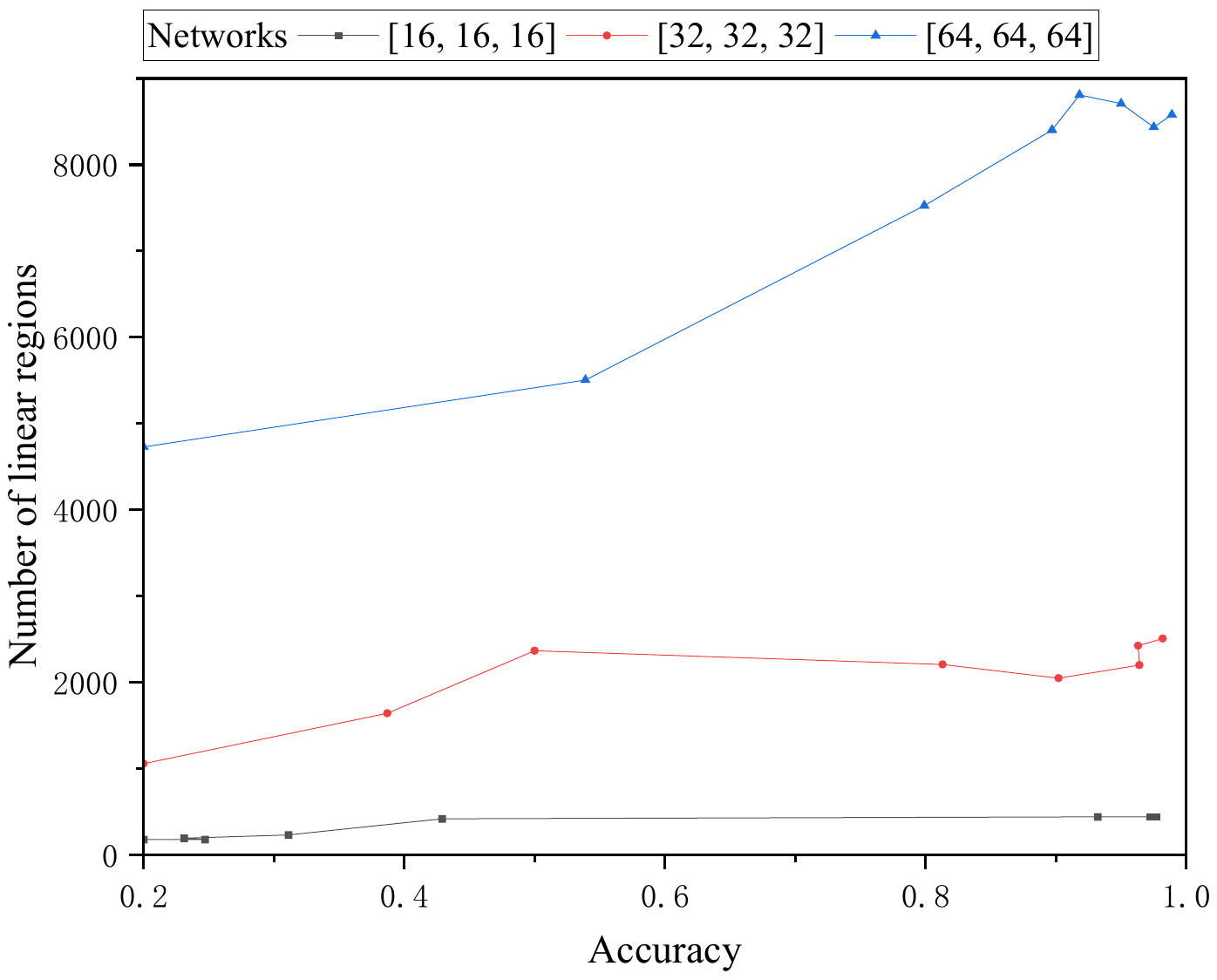}
	\end{minipage}
	\caption{The relationship between the number of linear regions, accuracy and epochs in different networks for the "$make$ $gaussian$ $quantiles$" dataset with 1000 samples.}
	\label{fig:7}
\end{figure}

We use three different networks, namely $[16, 16, 16]$, $[32, 32, 32]$ and $[64, 64, 64]$. Each network undergoes training for 100 epochs, and the number of linear regions is recorded. The experimental results for the "$make$ $moons$" dataset are presented in Fig. \ref{fig:6} and Table \ref{table 2}, while the experimental results for the "$make$ $gaussian$ $quantiles$" dataset are shown in Fig. \ref{fig:7} and Table \ref{table 3}.

The three networks exhibit strong memory capacity on both datasets, which differ from random data as they possess a regular structure. The visualization results of decision boundaries for networks with varying numbers of neurons in different epochs are presented in Fig. \ref{fig:8} and Fig. \ref{fig:9} in Appendix \ref{apd:second}. In these figures, each color corresponds to a specific type of data, and the white regions can be considered as approximate decision boundaries of the networks in their current states. Our empirical findings indicate that during the training process of ReLU networks, the linear regions are continuously optimized until they effectively capture the distribution of the data.

\begin{table}[htp]
	\centering
	\caption{The number of linear regions in different networks and epochs for the "$make$ $moons$" dataset with 1000 samples.}
	\setlength{\tabcolsep}{3.3mm}{
		\begin{tabular}{l|cccccccc}
			\toprule
			\diagbox [width=8em,trim=l] {DNNs} {Epoch} & 0 & 2 & 5  & 8 & 10 & 20 & 50 & 100  \\
			\hline
			$[16, 16, 16]$ & 206 & 198 & 319 & 358 & 363 & 395 & 456 & 467  \\
			$[32, 32, 32]$ & 1151 & 1151 & 1222 & 1401 & 1274 & 1533 & 1756 & 1725  \\
			$[64, 64, 64]$ & 4112 & 3897 & 3803 & 4472 & 5172 & 6354 & 6037 & 7082  \\
			\bottomrule
	\end{tabular}}\vspace{0cm}
	\label{table 2}
\end{table}

\begin{table}[H]
	\centering
	\caption{The number of linear regions in different networks and epochs for the "$make$ $gaussian$ $quantiles$" dataset with 1000 samples.}
	\setlength{\tabcolsep}{3.3mm}{
		\begin{tabular}{l|cccccccc}
			\toprule
			\diagbox [width=8em,trim=l] {DNNs} {Epoch} & 0 & 2 & 5  & 8 & 10 & 20 & 50 & 100  \\
			\hline
			$[16, 16, 16]$ & 207 & 174 & 189 & 228 & 415 & 437 & 438 & 436  \\
			$[32, 32, 32]$ & 1154 & 1638 & 2362 & 2204 & 2045 & 2197 & 2420 & 2505  \\
			$[64, 64, 64]$ & 4113 & 5501 & 7524 & 8402 & 8809 & 8706 & 8433 & 8579  \\
			\bottomrule
	\end{tabular}}\vspace{0cm}
	\label{table 3}
\end{table}

\section{Conclusion}
\label{5}
In this paper, we have investigated the relationship between linear regions and input distributions in networks based on the piecewise linear activation function ReLU. By computing the number of linear regions in the input space partition of ReLU networks, we have empirically found that complex inputs inhibit the network's ability to express linear regions. In particular, we have demonstrated the existence of a lower bound on the number of neurons required to express any one-dimensional input. Additionally, our two-dimensional visualization results reveal the iterative optimization process of decision boundaries during the training of fully connected ReLU networks. We plan to analyze the relationship between linear regions and input distributions in networks with other architectures in the future.

\bibliographystyle{unsrt}  
\bibliography{references}

\appendix

\section{Proof of Theorem \ref{(32)}}\label{apd:first}
Before proving Theorem \ref{(32)}, we need the conclusion of Theorem \ref{(33)} from \cite{book22}.

\begin{theorem}(adapted from \cite{book22}).
	\label{(33)}
Let $L$ be a ReLU network with one-dimensional input and output. We assume that the weights and biases are randomly initialized so that the pre-activation $o(x)$ of each neuron $q$ has a bounded average gradient.
\begin{equation}
	\label{(24)}
	\mathbb{E}[\|\nabla o(x)\|] \leq K, \quad\quad\quad  some \quad K>0 ,
\end{equation}
(\ref{(24)}) satisfies the independent zero-center weights initialization of the ReLU neural networks with variance as follows:
\begin{equation}
	\label{(25)}
	V=\frac{2}{\text {fan-in}}.
\end{equation}
Referring to \cite{book23}, during ReLU network initialization, each $q$ satisfies
\begin{equation}
	\label{(26)}
	\mathbb{E}\left[\|\nabla o(x)\|^{2}\right]=2.
\end{equation}
According to the above, suppose that an input subset $S$ of the ReLU network with $j$ neurons, and the average number of linear regions of $S$ is $P$, we have
\begin{equation}
	\label{(27)}
	\mathbb{E}[P] \approx \lvert S \rvert  \cdot j \cdot c, \quad S \subset \mathbb{R},
\end{equation}
where $c$ is the number of breakpoints, for ReLU networks, $c$ is equal to 1. (\ref{(27)}) states that $P$ is proportional to $j$. This result is also sufficient to calculate the number of linear regions along any fixed one-dimensional curve in any high-dimensional input. 
\end{theorem}
Considering the effect of random weights and biases on each neuron, for any reasonable initialization, suppose that the $o(x)$ of $q$ satisfies
\begin{equation}
	\label{(28)}
	\lvert o^{\prime}(x)\rvert=\gamma (1),
\end{equation}
referring to (\ref{(17)}), $x \mapsto o(x)$ cannot be highly oscillatory over a large part of $q$ with the input of $s_{i}$. Therefore, let the bias be $b_{q}$ and we have
\begin{equation}
	\label{(29)}
	\mathbb{E}[\#\{\text { solutions in } \left \{ o(x)=b_{q} \right \} \}]=Z(1) ,
\end{equation}
where $Z(1)$ is the solution of the equation we expect. Then, we expect each neuron to generate a fixed quantity of additional linear convex regions. 

Assuming that the minimum number of linear regions expressing $s_{i}$ is $Q$, we obtain the minimum number of neurons $q$ as follows:

(1) For independent zero-center weights initialization, we have
\begin{equation}
	\label{(30)}
	q\approx \frac{\mathbb{E}[Q]}{\lvert s_{i} \rvert \cdot c},\quad s_{i}\in S,
\end{equation}
where $c$ is the number of breakpoints, for ReLU DNNs, $c$ is equal to 1.

(2) Considering the effect of random weights and biases on each neuron, for any reasonable initialization, we have
\begin{equation}
	\label{(31)}
	q\approx \frac{\mathbb{E}[Q]- I}{\lvert s_{i} \rvert \cdot c}, \quad s_{i}\in S,
\end{equation}
where $c$ is the number of breakpoints, for ReLU DNNs, $c$ is equal to 1, and $I$ is the quantity of additional linear convex regions generated by all neurons. 

Therefore, as we augment the quantity of neurons until the linear regions generated by
these neurons are sufficient to represent the one-dimensional input $S$, we obtain the set of neuron quantities capable of representing $s_{i}$ as indicated in (\ref{(19)}) and (\ref{(23)}).
\clearpage
\section{Visualize the process of fitting decision boundaries}\label{apd:second}
\begin{figure}[htp]
	\centering
	\subfigure[]{\includegraphics[angle=0,width=1\textwidth]{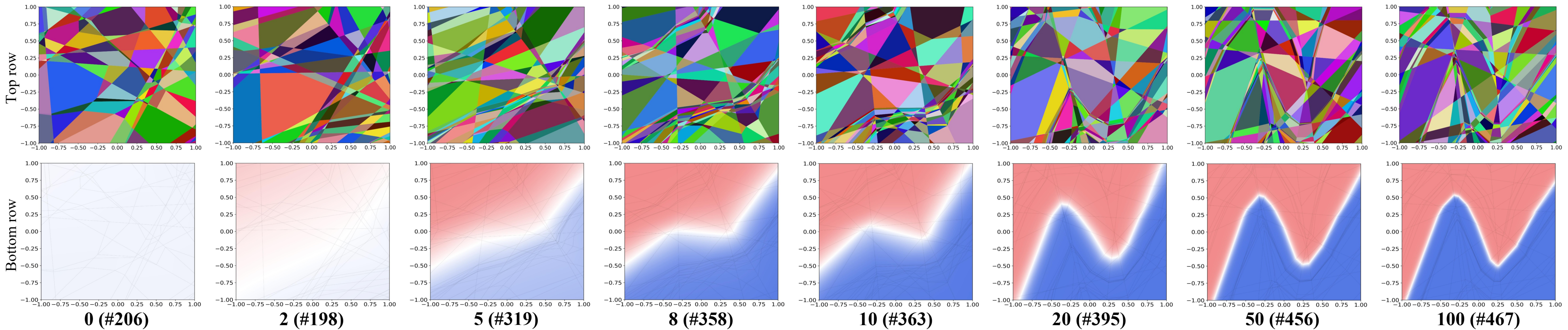}}
	\subfigure[]{\includegraphics[angle=0,width=1\textwidth]{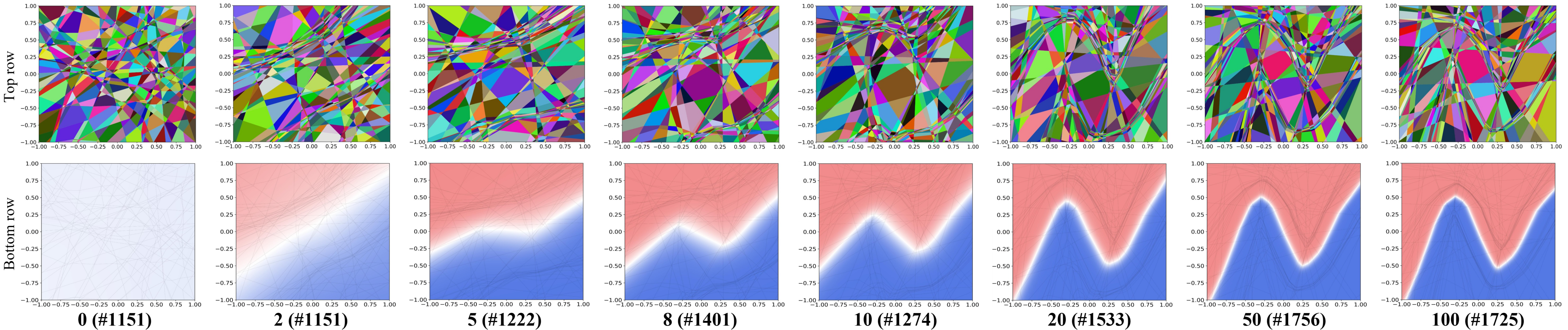}}
	\subfigure[]{\includegraphics[angle=0,width=1\textwidth]{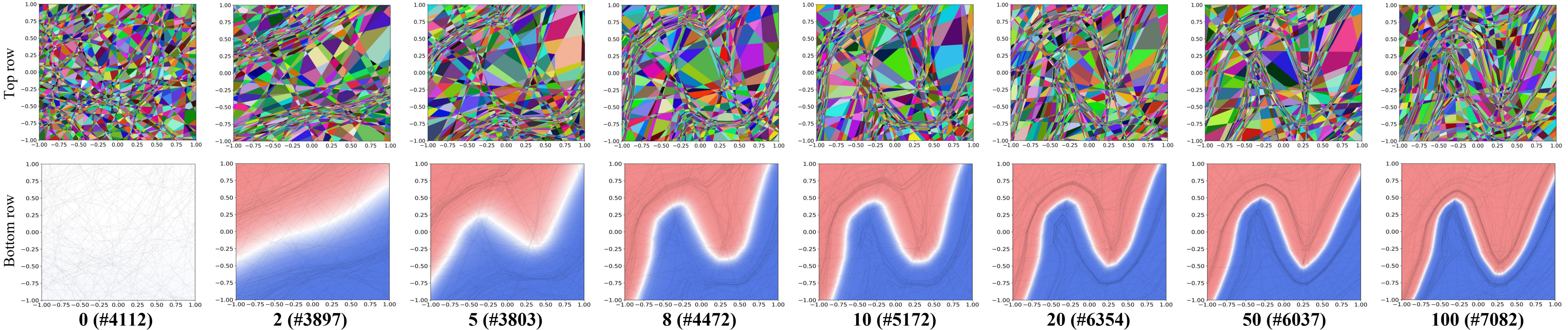}}
	\caption{The visualization results and the number of linear regions for different networks trained on the "$make$ $moons$" at $k$ epochs, for $k=0, 2, 5, 8, 10, 20, 50, 100$. Top row: Visualization of the linear regions. Bottom row: Visualization of decision boundaries, where each color represents each type of the input data, and the white regions can be approximately regarded as the decision boundaries for the current state of the network. (a) The network $[16, 16, 16]$. (b) The network $[32, 32, 32]$. (c) The network $[64, 64, 64]$.}
	\label{fig:8}
\end{figure}
\clearpage
\begin{figure}[H]
	\centering
	\subfigure[]{\includegraphics[angle=0,width=1\textwidth]{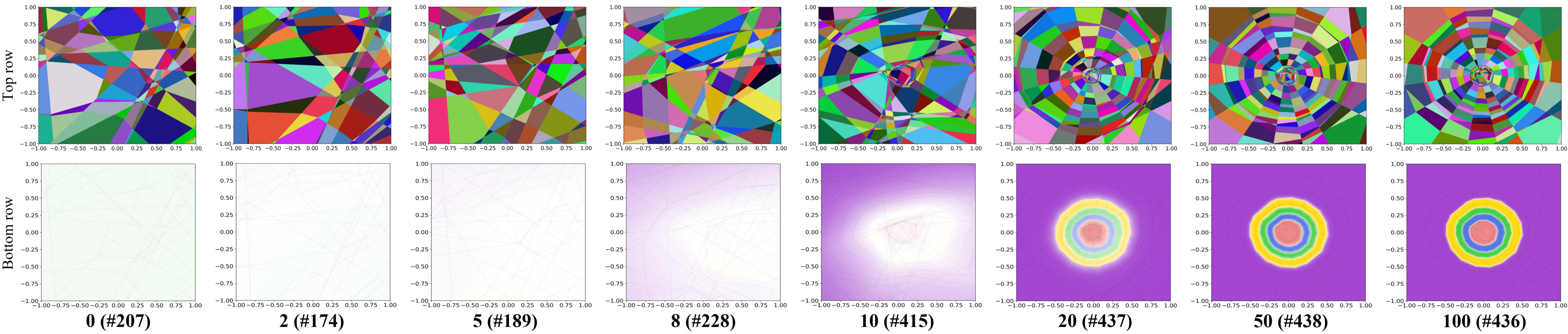}}
	\subfigure[]{\includegraphics[angle=0,width=1\textwidth]{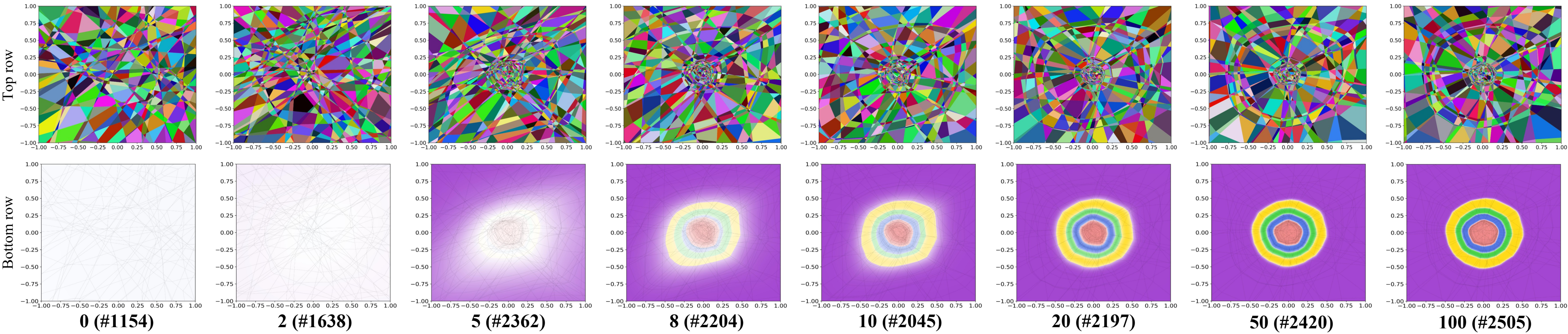}}
	\subfigure[]{\includegraphics[angle=0,width=1\textwidth]{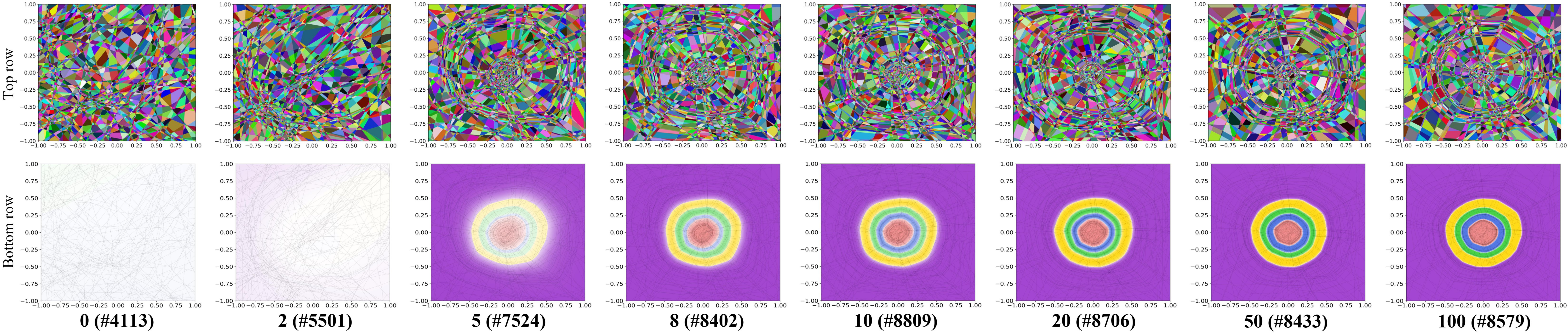}}
	\caption{The visualization results and the number of linear regions for different networks trained on the "$make$ $gaussian$ $quantiles$" at $k$ epochs, for $k=0, 2, 5, 8, 10, 20, 50, 100$. Top row: Visualization of the linear regions. Bottom row: Visualization of decision boundaries, where each color represents each type of the input data, and the white regions can be approximately regarded as the decision boundaries for the current state of the network. (a) The network $[16, 16, 16]$. (b) The network $[32, 32, 32]$. (c) The network $[64, 64, 64]$.}
	\label{fig:9}
\end{figure}

\end{document}